\documentclass[letterpaper, 10 pt, conference]{ieeeconf}  

\IEEEoverridecommandlockouts                              

\usepackage{xcolor}
\usepackage{comment}
\usepackage{paralist}
\usepackage{hyperref}
\usepackage{graphicx}
\usepackage{amsfonts}
\usepackage{amsmath}
\usepackage[english]{babel}
\usepackage{booktabs}
\usepackage{multirow}

\usepackage{subcaption}
\usepackage{floatrow}
\usepackage{sidecap}
\usepackage[affil-it]{authblk}

\title{Learning Sim-to-Real Dense Object Descriptors for Robotic Manipulation}

\author[1]{Hoang-Giang Cao}
\author[1,2]{Weihao Zeng}
\author[1,3]{I-Chen Wu$^\dagger$}
\affil[1]{Department of Computer Science, National Yang Ming Chiao Tung University, Taiwan}
\affil[2]{School of Computer Science, Carnegie Mellon University, United States}
\affil[3]{Research Center for IT Innovation, Academia Sinica, Taiwan}

\begin{document}
\maketitle
\def\thefootnote{$\dagger$}\footnotetext{Correspondence.}\def\thefootnote{\arabic{footnote}}
\def\thefootnote{}\footnotetext{Code and data: \url{https://github.com/hgiangcao/SRDONs}}\def\thefootnote{\arabic{footnote}}
.
\thispagestyle{empty}
\pagestyle{empty}


\begin{abstract}
It is crucial to address the following issues for ubiquitous robotics manipulation applications: 
(a) vision-based manipulation tasks require the robot to visually learn and understand the object with rich information like dense object descriptors; and
(b) sim-to-real transfer in robotics aims to close the gap between simulated and real data.
In this paper, we present Sim-to-Real Dense Object Nets (SRDONs), a dense object descriptors that not only understands the object via appropriate representation but also maps simulated and real data to a unified feature space with pixel consistency.
We proposed an object-to-object matching method for image pairs from different scenes and different domains.
This method helps reduce the effort of training data from real-world by taking advantage of public datasets, such as GraspNet.
With sim-to-real object representation consistency, our SRDONs can serve as a building block for a variety of sim-to-real manipulation tasks. 
We demonstrate in experiments that pre-trained SRDONs significantly improve performances on unseen objects and unseen visual environments for various robotic tasks with zero real-world training.

\end{abstract}


\section{Introduction}

Vision-based robotics reinforcement learning methods have enabled solving complex robotics manipulation tasks in an end-to-end fashion \cite{akkaya2019solving,khazatsky2021can,kalashnikov2021mt}.
The ability to understand unseen objects is one of crucial issues for robotics tasks. 
Although object segmentation is helpful, object-level segmentation ignores the rich structures within objects \cite{florencemanuelli2018dense}.
A better object-centric descriptor is critical for ubiquitous robotics manipulation applications.
Dense Object Nets (DONs) can learn object representation useful for robotics manipulation in a self-supervision manner \cite{florencemanuelli2018dense}.
The learned dense descriptors enabled interesting robotics applications, such as soft body manipulation and pick-and-place from demonstrations \cite{Chai2019,riya2020learningrope}.
DONs are trained with matching and non-matching pixel coordinates in pairs of images. 
However, in the data generation process in DONs, since each pair of individual training images comes from the same object configuration, this makes it hard to be used in different object configurations. 
Object configuration is the setup of the object's positions in a scene.
Thus, it becomes vital to learn explicitly from different object configurations for reliable object-centric descriptors.

\begin{figure}[t]
    \includegraphics[width=0.9\columnwidth]{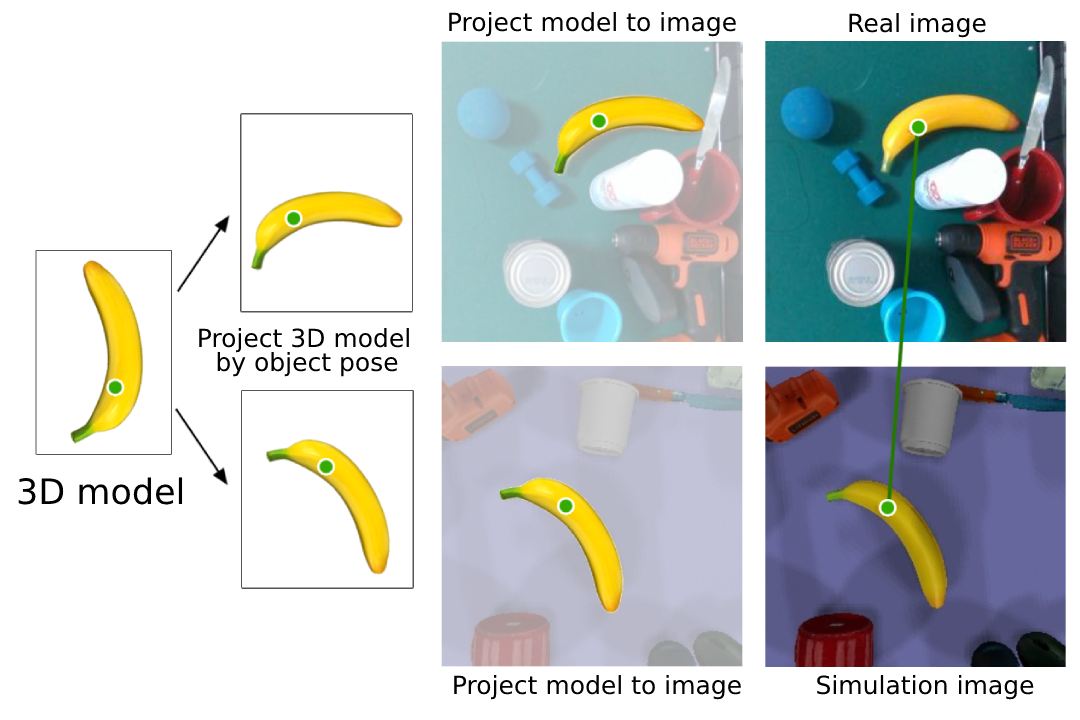}
    \caption{Object-to-object matching.}
    \label{fig:object_to_object_matching}
\end{figure}

Due to the difficulty of collecting a large amount of data in the real-world, which is crucial for learning-based robotics applications, we often train agents in the simulation and migrate to the real-world, but this poses the sim-to-real gap problem \cite{hofer2020perspectives}.
To successfully deploy learning-based robotics tasks into the real-world, a good dense object descriptors method needs not only to represent objects well but also represent simulation and real objects consistently.
Prior works focus on either solving the representation problem \cite{florencemanuelli2018dense} or the sim-to-real gap problem \cite{ho2020retinagan}.

In this paper, we present SRDONs (Sim-to-Real Dense Object Nets), a dense object descriptors with sim-to-real consistency.
To represent rich object structures with sim-to-real consistency, we utilize object poses and object meshes to automatically generate matching and non-matching pixel coordinates for image pairs from different object configurations (where the images come from different scenes) and different data domains (simulation or real-world).
Such data generation process enables training SRDONs using readily available large-scale public dataset.
SRDONs explicitly learn dense object descriptors from different object configurations in the simulation and the real-world. 
The resulting dense object descriptors can effectively represent simulation and real-world object information in a pixel consistent manner.
Furthermore, SRDONs exhibit great generalization ability from our experiments. And, SRDONs perform well on unseen objects in unseen visual environments in simulation and the real-world.

\textbf{Contributions.} The main contributions of this paper can be summarized as follows:
\begin{inparaenum}[1)]
    \item We present SRDONs, a dense object descriptors representation with sim-to-real consistency and generalization ability.
    \item We propose the matching pixel method for image pairs from different object configurations and different data domains, which helps reduce the effort of training data from real-world by taking advantage of public datasets, such as GraspNet \cite{fang2020graspnet}. 
    \item In experiments, we demonstrate the effectiveness of SRDONs in sim-to-real transfer on several robotics tasks and achieve high sim-to-real performances with unseen objects, unseen visual environments, and zero real-world training.
\end{inparaenum}

\section{Background}


\subsection{Dense Object Descriptors}

The ability to recognize and interact with unseen objects is critical in robotics applications. Recent works have explored unseen object segmentation \cite{xiang2020learning,xie2021unseen}.
However, the segmentation objective disregards the rich information within objects, e.g. curves and flat surfaces.
Dense object descriptors by \cite{florencemanuelli2018dense} is a promising direction for providing such ability.

Dense Object Nets (DONs) is a self-supervised method for generating dense object descriptors useful for robotics applications.
DONs learns from point to point correspondence, and are able to provide rich information within objects.
Recent works demonstrated the effectiveness of DONs in many challenging robotics applications.
The work in  \cite{riya2020learningrope} presented a method for manipulating ropes using dense object descriptors.
Another work by \cite{Chai2019} enforced an additional object-centric loss to enable multi-step pick-and-place from demonstration;
while in \cite{yang2021multiobject} imposed an additional object loss to achieve goal-conditioned grasping.
Cao {\em at el.} \cite{giang2021using} proposed Cluttered Object Descriptors (CODs) to represent the objects in a cluttered for robot picking cluttered objects.

\subsection{Sim-to-Real Transfer}

Training learning-based methods in the real-world are costly for robotics applications. Hence, bridging the gap between simulation and real-world is critical for ubiquitous learning-based robotics applications.
Recent works mainly focus on domain randomization and domain adaptation.

In domain randomization, we randomize the simulation so that the policies are robust enough to handle real-world data \cite{zhao2020sim}.
Such methods include randomizing textures, rendering, and scene configurations \cite{ho2020retinagan}. However, domain randomization requires careful task-specific engineering in selecting the kind of randomization.

In domain adaptation, we map simulation and real-world data into a knowledge preserving unified space.
We can directly map simulation images to real-world images, where GANs are commonly applied \cite{bousmalis2017unsupervised,kan2020rlcyclegan,ho2020retinagan}. However, GAN-based methods do not generalize well to unseen objects and scenes \cite{ho2020retinagan}, which is crucial for many robotics applications.
Prior works have also explored learning domain invariant features \cite{ganin2016domain,long2015learning,bousmalis2016domain}.
The work in \cite{jeong2020self} employed the temporal nature of robot experience. 
Another approach by \cite{ganin2016domain} separated task-specific and domain-specific knowledge via adversarial training.

While many of these prior works for dense object descriptors only used data in the same domain (only real data or simulation data) for training, we train the descriptor to match the pixels of the object between simulation and real-world images. 
Therefore, our SRDONs not only learn the useful object descriptors but also map the simulation and real-world images into a unified feature space with pixel consistency, addressing the sim-to-real problem.
	
\section{SRDONs: Sim-to-Real Dense Object Nets}


\subsection{Object-to-Object Matching}
\label{sec:object-to-object-matching}
Previous works \cite{florencemanuelli2018dense,Chai2019,giang2021using}, by using 3D TSDF (truncated signed distance function) reconstruction, can only generate matching pixels of static scenes in the same data domain.
Here, we proposed an object-to-object matching method with object poses and 3D models, which can generate matching points for images from different scenes and data domains.
Additionally, while other methods required collecting training data by running real robots, our method reduces time and cost by taking advantage of public datasets.

As illustrated in \autoref{fig:object_to_object_matching}, we compute the pixel coordinates corresponding to the 3D vertices on the same object model in each image to find matching points.
Suppose an image contains a set of objects, $O = (O_1,O_2,O_3, \ldots,O_n)$, and pose annotations for each object, $\Phi = (\Phi_1,\Phi_2,\Phi_3, \ldots,\Phi_n)$. 
The 3D model $O_i$ is given by a set of vertices $\textbf{V}_\textbf{i} = ( \textbf{X}, \textbf{Y}, \textbf{Z})^T$. 
To associate 3D model vertices with 2D pixel coordinates, we also need the projection matrix  $\mathcal{P}_i$ for each object $O_i$, where $\mathcal{P}_i$ is computed from the intrinsic matrix, $K$, and extrinsic matrix, $E_i$. $K$ deals with the camera properties, and is known from the camera properties; $E_i$ represents the translation, $t_i$, and the orientation $R_i$ of object $O_i$ with respect to the camera. $E_i$ is computed from the object pose $\Phi_i$. 
Note the pose annotations are in the camera frame, so we do not need to consider camera transformations.

The projection matrix $\mathcal{P}_i$ is:
\begin{equation} 
\centering
\mathcal{P}_i = K E_i  \quad \textrm{with}  \quad E_i =[R_i | t_i]
\end{equation}
We then project all 3D vertices $\textbf{V}_\textbf{i}$ of the object $O_i$ onto the image coordinate system to get their 2D corresponding pixel coordinates $(\textbf{u},\textbf{v})^T$.

\begin{equation}
\centering
\begin{bmatrix}
\textbf{u} \\\textbf{v} \\\textbf{1}
\end{bmatrix}= \mathcal{P}_i \begin{bmatrix}
 \textbf{X}' \\ \textbf{Y}' \\ \textbf{Z}' \\ \textbf{1}
\end{bmatrix}
\quad \textrm{with}  \quad
\begin{bmatrix}
\textbf{X}' \\ \textbf{Y}' \\ \textbf{Z}'
\end{bmatrix} = \Phi_i \begin{bmatrix}
\textbf{X} \\ \textbf{Y} \\ \textbf{Z}
\end{bmatrix}
\end{equation}




\begin{table}[hbt!]
\centering
\begin{tabular}{|c|c|c|c|}
\hline
Method &
  \begin{tabular}[c]{@{}c@{}}Robotic\\ sampling\end{tabular} &
  \begin{tabular}[c]{@{}c@{}}Multi-objects\\ (classes)\end{tabular} &
  \begin{tabular}[c]{@{}c@{}}Sim\\ to\\ Real\end{tabular} \\ \hline
\begin{tabular}[c]{@{}c@{}}Original DONs\cite{florencemanuelli2018dense}\end{tabular}     & Yes & Yes (3)  & No  \\ \hline
\begin{tabular}[c]{@{}c@{}}MCDONs\cite{Chai2019}\end{tabular}  & Yes & Yes (8)  & No  \\ \hline
\begin{tabular}[c]{@{}c@{}}LE DONs\cite{andras2021supervised}\end{tabular}           & Yes & No       & No  \\ \hline
\begin{tabular}[c]{@{}c@{}}MODONs\cite{yang2021multiobject}\end{tabular} & No  & Yes (16) & No  \\ \hline
\begin{tabular}[c]{@{}c@{}}Our SRDONs\end{tabular}        & No  & Yes (28) & Yes \\ \hline
\end{tabular}
\caption{Comparison of different datasets.}
\label{tab:s2r_dons_compare}
\end{table}

For a pair of images both containing some objects, we randomly sample from the object models a subset of their 3D vertices and calculate their corresponding 2D pixel coordinates in each image. 
The pixel coordinates in each image corresponding to a 3D model vertex are considered as matching pixel coordinates.
To deal with occlusion, we assign the pixel coordinate to the vertex closest to the camera in Euclidean distance.
Object-to-object matching enables generating matching in a variety of scenarios: different scene matching (dynamic scenes), sim-to-real matching, and multiple matching (finding matching between one object in an image with multiple of the same objects in another).

\subsection{Contrastive Loss}

We employ the contrastive loss from \cite{florencemanuelli2018dense} to enable self-supervised learning for SRDONs. 
Given an image $I \in \mathbb{R}^{W \times H \times d}$ where $d$ can be either $3, 4$ depending on whether the input is RGB or RGBD, we map $I$ to a dense descriptor space $\mathbb{R}^{W \times H \times D}$. 
Each pixel in $I$ has a corresponding $D$-dimensional feature vector.
Given a pair of images, the matching pixels coordinates, and the non-matching pixels coordinates, we optimize the dense descriptor network, $f$, to minimize the L2 distances between descriptors of matching pixels, and keep descriptors of non-matching pixels $M$ distance apart.
\begin{equation}
\mathcal{L}_{\textrm{\small{m}}}(I_a, I_b) = \frac{1}{N_{\textrm{\small{m}}}} \sum\limits_{N_{\textrm{\footnotesize{m}}}}  \lVert f(I_a)(u_a) - f(I_b)(u_b) \rVert^2_2
\end{equation}
\begin{equation}
\label{eq:non-match}
\resizebox{\hsize}{!}{
$
\mathcal{L}_{\textrm{\small{nm}}}(I_a, I_b) = \frac{1}{N_{\textrm{\small{strict\_nm}}}} \sum\limits_{N_{\textrm{\footnotesize{nm}}}}  max(0, M - \lVert f(I_a)(u_a) - f(I_b)(u_b) \rVert_2)^2
$}
\end{equation}
\begin{equation}
\mathcal{L}(I_a, I_b) = \mathcal{L}_{\textrm{\small{m}}}(I_a, I_b) + \mathcal{L}_{\textrm{\small{nm}}}(I_a, I_b)
\end{equation}
where "m" is matching, and "nm" is non-matching; $N_{\textrm{m}}$ is the number of matches, and $N_{\textrm{nm}}$ is the number of pairs of non-matching pixels ; $f(I)(u)$ is the descriptor of $I$ at pixel coordinate $u$; $N_{\textrm{strict\_nm}}$ is the number of pairs of non-matching descriptors within $M$ distance to each other, namely the number of non-zero terms in the summation term in \autoref{eq:non-match}.

\subsection{Data Collection and Training SRDONs}
Previous works \cite{florencemanuelli2018dense,Chai2019} required the use of a real robot arm to collect data.
In contrast, as described in Subsection \ref{sec:object-to-object-matching}, our proposed matching method enables to use real data from public datasets and simulated data generated from the simulation.
By this approach, we not only have easy access to diverse objects but also reduce the time and cost of real training data collection.
Table \ref{tab:s2r_dons_compare} compares our dataset with other works. 

\textbf{Real Data.} In this paper, we mainly use the real data from GraspNet \cite{fang2020graspnet}. 
The dataset provides 97,280 RGBD images of 88 objects over 190 cluttered scenes.
Each scene contains 9-10 objects placed at random positions on a tabletop. 
They capture 256 images per scene with different view poses and recorded the camera pose, the 6D pose of each object corresponding to each image.
However, to make the view of each scene sufficiently different, we downsampled the number of images to 50 images per scene. 

\textbf{Simulation Data.} We use V-REP simulator to generate the simulation image.
For each scene, we randomly drop 9 to 10 objects on a table.
We then use a camera to capture the RGBD images and record the camera poses and object poses with the same format as GraspNet dataset.
We also apply texture randomization and background randomization for generalization purpose.

\begin{figure}[t!]
  \centering
\includegraphics[width=0.90\columnwidth]{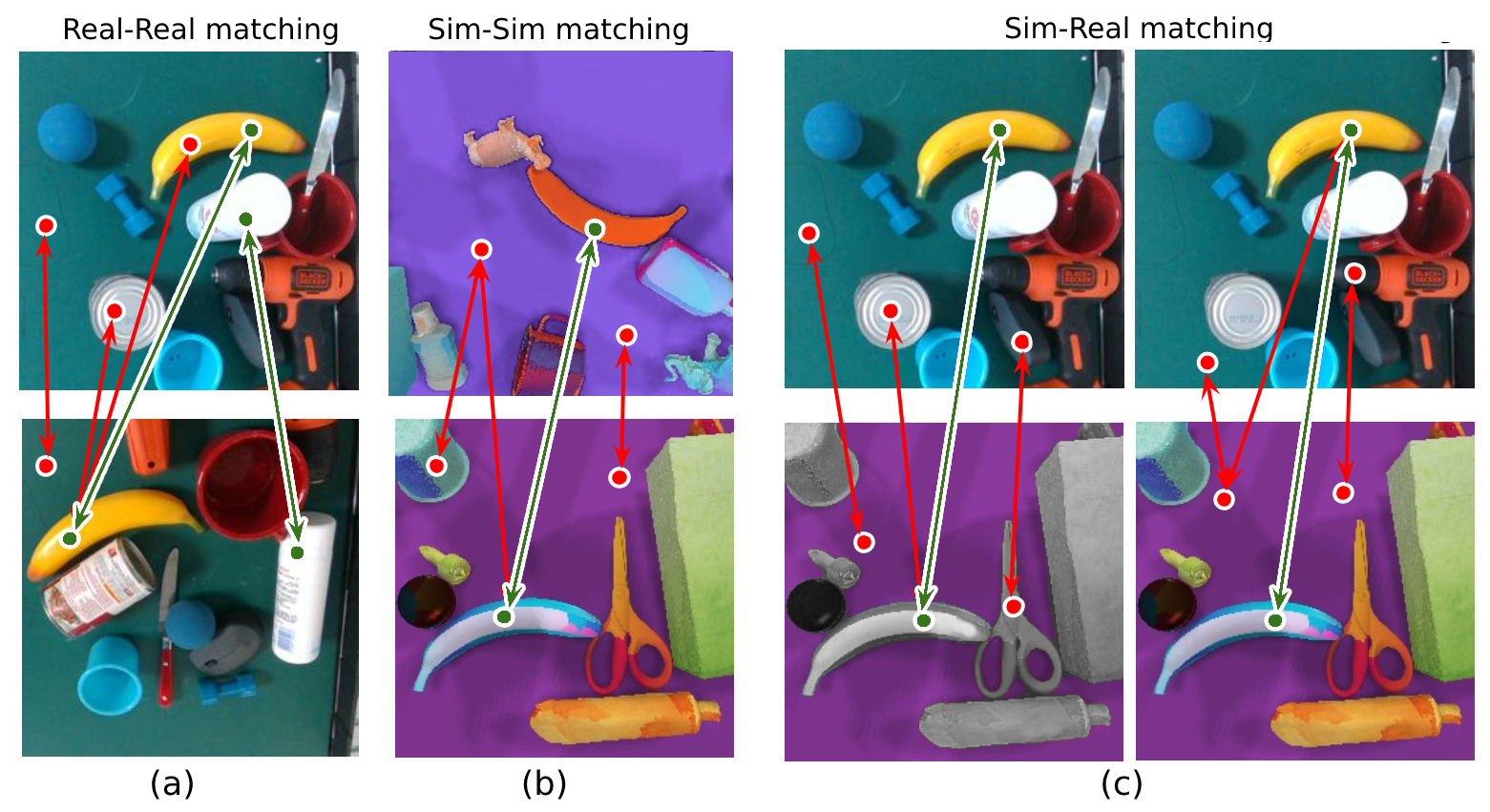}
  \caption{Different pairing types for training the descriptors.
  (a) Real-Real pairing.
  (b) Sim-Sim pairing.
  (c) Sim-Real pairing without (left) and with texture randomization (right).
  Green lines indicate pairs of match points; while red lines indicate pairs of non-match points.}
  \label{fig:matching_example}
\end{figure}

\textbf{Pairing Images.}
With the proposed object-to-object matching in \autoref{sec:object-to-object-matching}, we can generate matching pixel coordinates for any pair of images independent of data domain (simulation or real-world).
We have 3 different types of pairing:
\begin{inparaenum}[(a)]
    \item Sim-Sim: a pair of images is sampled from simulated images.
    \item Real-Real: a pair of images is sampled from real images.
    \item Sim-Real: one image is sampled from simulated images, and the other comes from the real images.
\end{inparaenum}
\autoref{fig:matching_example} shows some examples of different pairing types.

\textbf{Training.}
During each training step, we uniformly sample pairing types (Sim-Sim, Real-Real, and Sim-Real). 
Once a type has been sampled, we then choose whether the two images are from the same scene or different scenes (with the probability of 30\% and 70\%, respectively). 
For Sim-Sim and Real-Real matching, two images may come from the same or different scenes, while for Sim-Real matching, two images have to come from different scenes, since they come from different data domains. 
For each pair of images, we sample 1000 pairs of matching points, and 5000 pairs of non-matching points (object to object, object to background, background to background).
More details about collecting data and training are provided in the\textit{ accompanying video}.

\begin{figure}[t]
\includegraphics[width=0.9\linewidth]{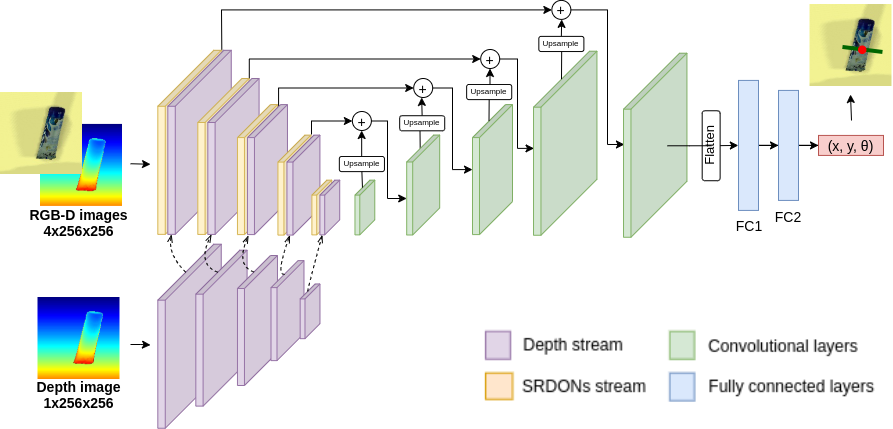}
\caption{The architecture of using SRDONs in grasping tasks.}
\label{fig:supervised_learning_network_structure}
\end{figure}

\begin{table*}[!h]
\begin{tabular}{|l|c|c|c|c|c|}
\hline
\multicolumn{1}{|c|}{} &
  Input &
  \begin{tabular}[c]{@{}c@{}}Sim-Sim\\ Rd-100\end{tabular} &
  Real-Real &
  \begin{tabular}[c]{@{}c@{}}Sim-Real\\ Rd-0\end{tabular} &
  \begin{tabular}[c]{@{}c@{}}Sim-Real\\ Rd-100\end{tabular} \\ \hline
Original DONs\cite{florencemanuelli2018dense}     & RGB  & 0.155 / 0.265          & 0.430 / 0.203          & 0.143 / 0.281          & 0.140 / 0.289          \\ \hline
MODONs\cite{yang2021multiobject}            & RGB  & 0.157 / 0.267          & \textbf{0.947 / 0.027} & 0.189 / 0.264          & 0.164 / 0.279          \\ \hline
CODs \cite{giang2021using}             & RGBD & \textbf{0.939 / 0.063} & 0.256 / 0.238          & 0.424 / 0.216          & 0.389 / 0.230          \\ \hline
SRDONs - Rd-0   & RGB  & 0.248 / 0.241          & 0.935 / 0.063          & 0.695 / 0.140          & 0.258 / 0.264          \\ \hline
SRDONs - Rd-80  & RGB  & 0.857 / 0.095          & 0.910 / 0.085          & 0.915 / 0.092          & 0.908 / 0.098          \\ \hline
SRDONs - Rd-100 & RGB  & 0.899 / 0.082          & 0.904 / 0.084          & 0.915 / 0.086          & 0.936 / 0.088          \\ \hline
SRDONs - Rd-0   & RGBD & 0.248 / 0.248          & 0.941 / 0.059          & 0.684 / 0.140           & 0.247 / 0.269          \\ \hline
SRDONs - Rd-80  & RGBD & 0.908 / 0.079          & 0.917 / 0.082          & \textbf{0.925 / 0.084} & \textbf{0.941 / 0.086} \\ \hline
SRDONs - Rd-100 & RGBD & 0.911 / 0.077          & 0.913 / 0.076          & 0.911 / 0.089          & 0.933 / 0.090          \\ \hline
\end{tabular}
\caption{Evaluate matching results with different methods (in accuracy/error distance).
For fairness, we use the same descriptor dimension (D=8) for all models.
The above results are also visualized as in the \textit{accompanying video}.
}
\label{tab:srdons_tranining_result}
\end{table*}


\subsection{SRDONs for Robotics Learning Tasks}
\label{sec:policy_networ}
We want to use SRDONs to serves as a building block for robotic tasks. 
The work in \cite{giang2021using} proposed a network structure that can use the intermediate layers of the descriptor network for training a reinforcement learning task. 
We adopt their method, and extend to supervised learning method.

To use SRDONs for training the reinforcement learning (RL) task, we simply apply the same structure as proposed in \cite{giang2021using} (further details in the\textit{ accompanying video}), which is based on actor-critic RL, namely PPO.
In the supervised learning task, we slightly modify the network structure. 
We first remove the critic head, and then, change the actor head which originally is fully convolutional layers to fully connected layers. 
Specifically, in our experiment, we use the grasping task to demonstrate the use of SRDONs with supervised learning. 
Grasping task requires the agent to predict the grasping position ($x$, $y$) and the top-down grasping pose ($\theta$) to grasp the object. 
Therefore, the output of the network in this task is the pose ($x$, $y$, $\theta$) $\in \mathbb{R}^3 $.
\autoref{fig:supervised_learning_network_structure} shows how we combine the intermediate layer of SRDONs and the depth stream in a U-Net structure.

\section{Experimental Results}
\label{sec:experiments}

We conduct experiments to evaluate SRDONs performances on providing good sim-to-real object descriptors in Subsection \ref{sec:srdon_evaluation}.
Then we use a pre-trained SRDONs as the building block for solving two robotic manipulation tasks: two-fingered grasping in Subsection \ref{sec:srdons_grasping} and picking cluttered objects with suction in Subsection \ref{sec:srdons_picking}.
The robotic tasks are both trained entirely in the simulation, and directly tested in the real-world with zero real-world training.


\subsection{Evaluation of Sim-to-Real Object Descriptors}
\label{sec:srdon_evaluation}
We evaluate the performance of descriptors by finding matching interest points, as in \cite{florencemanuelli2018dense}.
We employ two evaluation metrics: the accuracy of matching correct objects, and the matching error distance normalized by image diagonal distance. 
Given a pair of source and target images and a point $p$ on the source image that belongs to an object $O$, let $p^*$ indicate the true match point in the target image (disregarding the case of occlusion as described above), and $p'$ indicate the best match point by using a given descriptor.
For the former, the object matching accuracy of matching the correct objects, i.e., $p'$ and $p^*$ are on the same object in the target image. 
For the latter, the matching error distance is the average distance between $p^*$ and $p'$.

In the experiments, we use the following methods as the baselines: 
\begin{inparaenum}[(a)]
\item The original DONs \cite{florencemanuelli2018dense}.
\item The Multi-Object DONs (MODONs) \cite{yang2021multiobject}.
\item The Cluttered Object Descriptors (CODs) \cite{giang2021using}.
\end{inparaenum}
When training the original Dense Object Nets, we use Real-Real in the same scenes; for the Multi-Object DONs, we use Real-Real pairing in both the same and different scenes;
for the CODs, we use Sim-Sim in both the same and different scenes;
and our proposed SRDONs uses Sim-Real paring only.
We also evaluated the effects of randomizing object textures by randomizing 0\%, 80\%, and 100\% of the object textures when training SRDONs, denote as Rd-0, Rd-80, and Rd-100, respectively.
Additional training details and matching results are reported in the \textit{accompanying video}. 

\begin{figure}[h]
\begin{center}
  \includegraphics[width=0.95\textwidth]{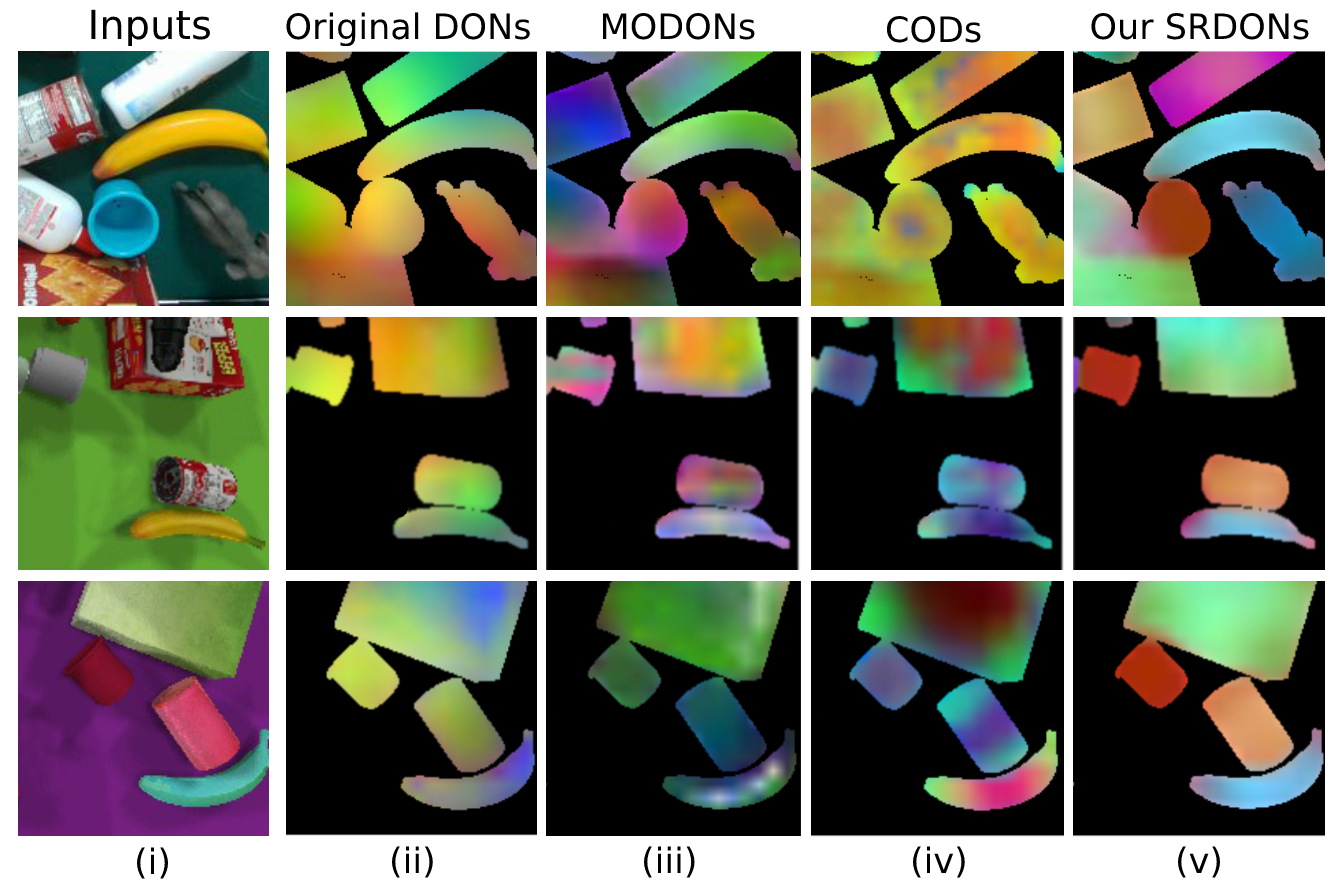}
  \caption{Evaluate sim-to-real descriptor translation.
  (i): different inputs: real image (top), simulated image without texture randomization (middle) and with texture randomization (bottom).
  (ii)-(v): descriptors generated by different methods
  Our SRDONs is able to represent the objects consistently in different inputs.
  The colors of these descriptors are produced in a similar way to t-SNE.}
 \label{fig:srdons_compare_description}
\end{center}  
\end{figure}

\textbf{Sim-to-Real Finding Matching Points.}
To evaluate the matching performance, we select 500 unseen image pairs for each type of pairing.
For each pair of images, we sample 1000 matching points and evaluate the matching performances. \autoref{tab:srdons_tranining_result} shows the experimental results of finding matching points of objects in the same domain (Sim-Sim Rd-100, Real-Real) and different domains (Sim-Real Rd-0 and Rd-100). 
(Note that Sim-Sim Rd-0 is less interesting so ignored in the table.)
We can see that the original DONs method (in the second row), which trained with same scenes only, fails to represent multi-object scenes.
In different domains like Sim-Real, our SRDONs shows the best performance in the rightmost two columns.
In the same domains like Sim-Sim and Real-Real, the result of our SRDONs are close to other methods which are trained with these specific types of paring, while our method used Sim-Real pairing only.
We visualize the descriptors of different methods in \autoref{fig:srdons_compare_description}, which shows that SRDONs can represent objects in simulation and real-world images with pixel consistency. 
Furthermore, texture randomization enables SRDONs to focus on the object geometry rather than color, as shown by the consistent object representation under texture randomization.

\begin{figure}[]
\begin{center}
  \begin{subfigure}[t]{0.49\linewidth}
        \includegraphics[width=\linewidth]{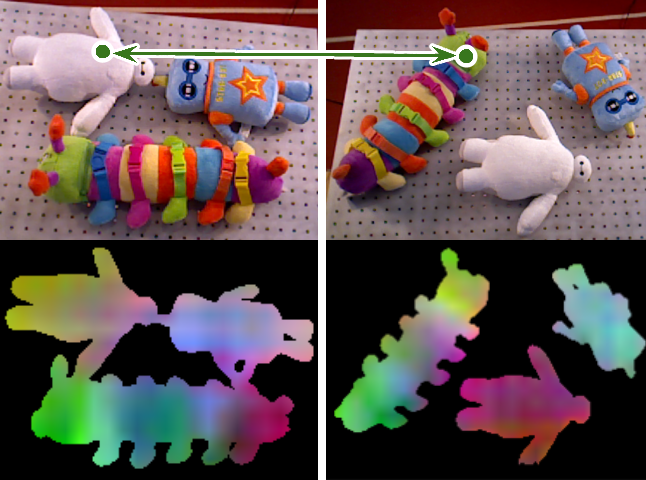}
        \caption{Original DONs - RGB.}
        \label{fig:don_unseen}
    \end{subfigure}
    \begin{subfigure}[t]{0.49\linewidth}
        \includegraphics[width=\linewidth]{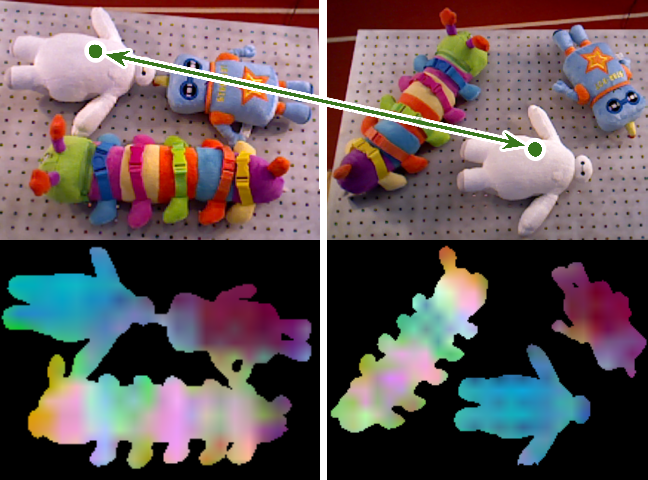}
        \caption{SRDONs - RGBD - Rd-80.}
        \label{fig:srdons_rd_80_unseen}
    \end{subfigure}
  \caption{
    Compare performances on \emph{unseen} objects between (a) the original DONs trained with RGB input and (b) SRDONs trained with RGBD input with Rd-80 setting.
    In each sub-figure, the top two images are inputs from different scenes, and the bottom two images are the corresponding descriptors of the above inputs.
    Green lines indicate match points based on the descriptors.
   }
  \label{fig:srdons_unseen}
\end{center}  
\end{figure}

For training the descriptors, we leverage public datasets like GraspNet, however, our SRDONs also works with unseen objects. \autoref{fig:srdons_unseen} shows the result of our SRDONs when performs testing on unseen objects.
We can see that DONs fails to represent and find the matching points with unseen objects in a multiple-object scene.
In contrast, our SRDONs is able to represent objects in the images consistently in the representation space and perform better matching.
Moreover, the matching performance of our method is improved by adding texture randomization and depth information, which are not considered in other DONs-based methods.

\begin{figure}[h!]
\begin{center}
  \begin{subfigure}[]{0.45\linewidth}
        \includegraphics[width=\linewidth]{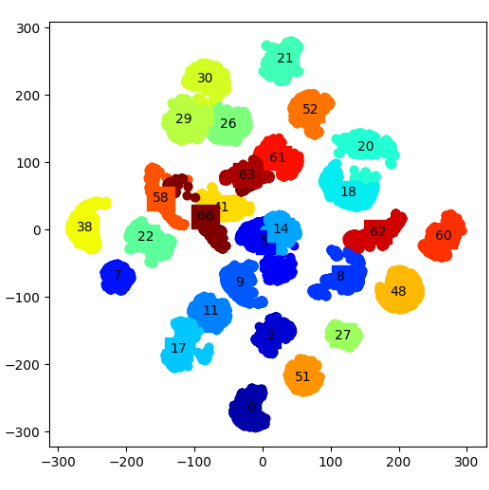}
        \caption{}
        \label{fig:tsne_real}
    \end{subfigure}
    \begin{subfigure}[]{0.45\linewidth}
        \includegraphics[width=\linewidth]{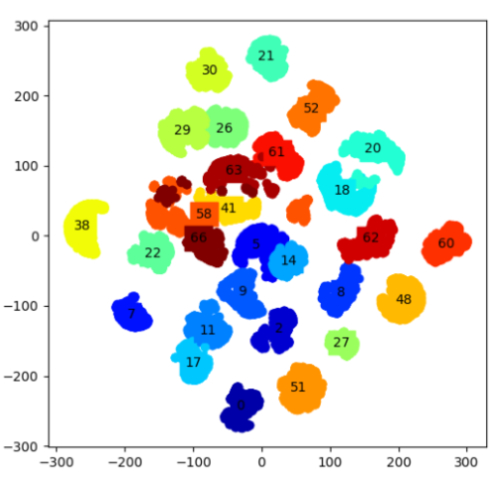}
        \caption{}
        \label{fig:tsne_sim}
    \end{subfigure}
  \caption{
  Clustering by t-SNE of object pixel-descriptors produced by SRDONs in simulation (a) and real (b) images.
  The points from the same object are marked in the same color.
  }
  \label{fig:srdons_compare_description_full_2}
\end{center}  
\end{figure}

\textbf{Sim-to-Real Multi-Object Consistent Evaluation.}
We conduct the experiment to verify that SRDONs are able to represent objects in simulation and real-world images with object consistency.
We use 100 unseen images from both simulation and real images. 
Each image contains 9 to 10 objects. 
We feed these images through the SRDONs, and randomly select 1000 pixel-descriptors per image.
Then, we use t-SNE to project the selected pixel-descriptors into two-dimensional for visualization, as shown in \autoref{fig:tsne_real} and \ref{fig:tsne_sim}.
In particular domain, the descriptors of the same objects are clearly distinguishable from the other objects.
While, in different domains, the descriptors of the same object from the real-world and the simulation reside in similar regions.

\subsection{Object Grasping}
\label{sec:srdons_grasping}
We used SRDONs as a building block in a robotics two-fingered grasping task to grasp an object that placed at randomly pose on the table.
We use the supervised learning (described in Subsection \ref{sec:policy_networ}) to predict the grasping pose, though applying reinforcement learning with continuous action spaces (DDPG \cite{Lillicrap2016ContinuousCW}) or discretizing the continuous action space will also work in this task.

In the simulation, we use Mean Square Error (MSE) between the ground-truth and the predicted grasping poses angle in radiance as the evaluation metrics in training and testing. 
In the real-world experiments, we do a real grasping by the robot and measure the success grasp rate.

We use the following methods as baselines: 
\begin{inparaenum}[(a)]
\item Domain randomization methods with RGB and RGBD as the inputs, denoted by RGB and RGBD respectively.
The model is similar to \autoref{fig:supervised_learning_network_structure}, however, we replace the SRDONs stream by RGB stream, which is a trainable ResNet34\_8s.
\item The model is similar to in \autoref{fig:supervised_learning_network_structure}, but we remove the SRDONs and using depth only, denoted by Depth.
\item The method proposed by \cite{giang2021using} for picking cluttered objects in simulation, denoted by CODs.
\item Our supervised learning method proposed in Subsection \ref{sec:policy_networ}, denoted by SRDONs.
\end{inparaenum}

In the simulation, we generated the ground truth grasping orientations, e.g., we place objects on the table in such a way that the grasping orientation is zero, where the robot can grasp the objects with zero z-rotation. 
We captured observations from different camera poses, and calculated the grasping location and the z-rotation ($x$,$y$,$\theta$) labels via camera poses.
For training data, we captured 50 RGBD images for each 28 objects from the GraspNet train slipt.
We then test the grasping methods with 15 objects from the GraspNet test splits.
In the real-world, we use a parallel gripper to grasp a single object that is placed on the table at random position and orientation. 
Each model performs 20 grasping trials (repeat twice with 10 objects collected from our lab). 

\autoref{tab:s2r_grasping} shows the results for grasping orientations prediction in both simulation and real-world environments.
In the simulation, SRDONs achieve the minimal average error of 0.15 radiance (8.59 degrees) on unseen objects.
Furthermore, the agent with SRDONs also outperformed others in the real-world by achieving 90\% success grasp rate without any further training. 
We can see that the models with depth information performed better than the others that use RGB input only.
Training details and real experiment videos are provided in the \textit{accompanying video}.
\begin{table}[H]
\centering
\begin{tabular}{|l|c|c|c|}
\hline
Method &
  \begin{tabular}[c]{@{}c@{}}Sim train\\ (radiance)\end{tabular} &
  \begin{tabular}[c]{@{}c@{}}Sim test\\ (radiance)\end{tabular} &
  \begin{tabular}[c]{@{}c@{}}Real\\ world\end{tabular} \\ \hline

\begin{tabular}[c]{@{}l@{}}RGB\end{tabular} &
  \begin{tabular}[c]{@{}c@{}}0.09\end{tabular} &
  \begin{tabular}[c]{@{}c@{}}0.49 \end{tabular} &
  65\% \\ \hline
\begin{tabular}[c]{@{}l@{}}RGBD\end{tabular} &
  \begin{tabular}[c]{@{}c@{}}0.061\end{tabular} &
  \begin{tabular}[c]{@{}c@{}}0.35\end{tabular} &
  85\% \\ \hline
  Depth &
  \begin{tabular}[c]{@{}c@{}}0.08\end{tabular} &
  \begin{tabular}[c]{@{}c@{}}0.36 \end{tabular} &
  85\% \\ \hline
  CODs &
  \begin{tabular}[c]{@{}c@{}}0.071 \end{tabular} &
  \begin{tabular}[c]{@{}c@{}}0.39 \end{tabular} &
  50\% \\ \hline
SRDONs &
  \textbf{\begin{tabular}[c]{@{}c@{}}0.035\end{tabular}} &
  \textbf{\begin{tabular}[c]{@{}c@{}}0.15\end{tabular}} &
  \textbf{90\%} \\ \hline
\end{tabular}
\caption{Result of grasping in simulation and real-world.}
\label{tab:s2r_grasping}
\end{table}


\subsection{Picking Cluttered General Objects}
\label{sec:srdons_picking}

Now, we used SRDONs as a building block for a more complex robotics picking task. 
Similarly as \cite{giang2021using}, we train an agent with reinforcement learning to pick cluttered objects with a suction pad. 
We have two metrics for evaluating the performance. 
The first is the rate of completion for all runs. A run is completion if all objects are picked before the episode terminates.
The second is the average number of objects picked in all runs.
In this picking cluttered objects task, we use the similar baselines to those in grasping task in Subsection \ref{sec:srdons_grasping}, but with reinforcement learning version.


In the simulation, we train each method in with 10 random objects sampled from GraspNet train split.
We then test with 20 and 30 objects from GraspNet test splits, and novel household objects.
In the real-world, we directly use the trained policy in the simulation to pick 10 novel household objects without any fine-tuning.


The experimental results in the simulation are shown in \autoref{tab:srdons_picking_completion} and \autoref{tab:srdons_picking_avg_pick_object}. 
Our method with SRDONs clearly out-performed other methods on all of the metrics, and are also efficient to be generalized to more cluttered scenarios with unseen objects.
When directly applying the trained policy in the simulation to the real-world testing, our method can successfully pick all 10 objects within 12.81 steps (78.1\% success pick rate), which is better than other methods (as shown in \autoref{tab:s2r_real_picking}).
Training details and real experiment videos are provided in the \textit{accompanying video}.

  \begin{table}[H]
\centering
\begin{tabular}{|l|c|c|c|}
\hline
\multicolumn{1}{|c|}{Dataset} &
  \begin{tabular}[c]{@{}c@{}}Grasp-\\ Net\end{tabular} &
  \begin{tabular}[c]{@{}c@{}}Grasp-\\ Net\end{tabular} &
  \begin{tabular}[c]{@{}c@{}}Novel\\ objects\end{tabular} \\ \hline
\multicolumn{1}{|c|}{\#obj} & 20      & 30      & 20      \\ \hline
RGB                         & 33.8\% & 24.5\% & 16.1\% \\ \hline
RGBD                        & 39.2\% & 25.1\% & 43.6\% \\ \hline
Depth                       & 89.2\% & 77.6\% & 68.3\% \\ \hline
CODs                        & 95.3\% & 92.9\% & 95.1\% \\ \hline
SRDONs                      & \textbf{97.8\%} & \textbf{94.1\%} &\textbf{ 97.5\%} \\ \hline
\end{tabular}
\caption{Picking completion rates in simulation.}

\label{tab:srdons_picking_completion}

\end{table}
\begin{table}[h]
\vspace*{-15pt}
\centering
\begin{tabular}{|l|c|c|c|}
\hline
\multicolumn{1}{|c|}{Dataset} &
  \begin{tabular}[c]{@{}c@{}}Grasp-\\ Net\end{tabular} &
  \begin{tabular}[c]{@{}c@{}}Grasp-\\ Net\end{tabular} &
  \begin{tabular}[c]{@{}c@{}}Novel\\ objects\end{tabular} \\ \hline
\multicolumn{1}{|c|}{\#obj} & 20    & 30    & 20    \\ \hline
RGB                         & 15.8 & 19.5 & 12.9 \\ \hline
RGBD                        & 16.9 & 23.4 & 16.1 \\ \hline
Depth                       & 18.9 & 25.5 & 18.1 \\ \hline
CODs                        & 19.1 & 28.3 & 18.9 \\ \hline
SRDONs                      & \textbf{19.7} & \textbf{29.6} & \textbf{19.2} \\ \hline
\end{tabular}
\caption{Average number of picked objects in simulation.}
\label{tab:srdons_picking_avg_pick_object}

\end{table}

\begin{table}[h]
\vspace*{-15pt}
\centering
\begin{tabular}{|l|c|c|c|}
\hline
Method &
  \begin{tabular}[c]{@{}c@{}}Completion\\ rate\end{tabular} &
  \begin{tabular}[c]{@{}c@{}}Success\\ rate\end{tabular} &
  \begin{tabular}[c]{@{}c@{}}Average\\ step\end{tabular} \\ \hline
\begin{tabular}[c]{@{}l@{}}RGB\end{tabular}  & 63.63\% & 61.12\% & 16.36 \\ \hline
\begin{tabular}[c]{@{}l@{}}RGBD\end{tabular} & 90.9\%  & 70.97\% & 14.90 \\ \hline
Depth                                                      & 72.72\% & 62.15\% & 16.09 \\ \hline
CODs                                                            & 72.72\% & 61.8\% & 16.18 \\ \hline
SRDONs                                                          &\textbf{ 100\%}   &\textbf{ 78.1\%}  & \textbf{12.81} \\ \hline
\end{tabular}
\caption{Result of picking objects in real-world.}
\label{tab:s2r_real_picking}
\end{table}


\section{Conclusion}
\label{sec:conclusion}
This paper presents SRDONs, a dense object descriptors representation with sim-to-real consistency.
Our method addresses both of the object representation problem and the sim-to-real gap problem. 
Through experiments, we demonstrated that our method can provide useful object information while representing simulation and real-world objects with pixel consistency. 
We showed that SRDONs enabled zero-shot sim-to-real transfer in robotic manipulation tasks on unseen objects and unseen visual environments.
With the representation power of SRDONs, we expect to accelerate the sim-to-real deployment process for robotics applications.

\clearpage
\bibliographystyle{plain}
\bibliography{root}

\end{document}